\pdfoutput=1
\documentclass[11pt,a4paper]{article}
\usepackage[dvipsnames]{xcolor}
\usepackage[hyperref]{acl2021}
\usepackage{times}
\usepackage{latexsym}
\usepackage{inconsolata}
\usepackage{microtype}
\usepackage[fleqn]{amsmath}
\usepackage{amsfonts}
\usepackage{url}
\usepackage{booktabs}
\usepackage{multirow}
\usepackage{xspace}
\usepackage{bm}
\usepackage{tikz}
\usepackage{tikz-dependency}
\usepackage{tabularx}
\usepackage[T5,T1]{fontenc}
\usepackage[utf8]{inputenc}
\usepackage[title]{appendix}
\usepackage{fancybox}
\usepackage{amsthm}
\usepackage{mathtools}
\usepackage{amssymb}
\usepackage{pgfplots, pgfplotstable}
\usepackage{filecontents}

\usetikzlibrary{shapes.misc, fit, positioning, calc}

\tikzset{
    vertex/.style={circle,draw,minimum size=1.5em},
    edge/.style={->,> = latex'}
}

\aclfinalcopy %
\def\aclpaperid{***} %

\title{
  TGIF: \underline{T}ree-\underline{G}raph \underline{I}ntegrated-\underline{F}ormat Parser for Enhanced UD\\
  with \underline{T}wo-Stage \underline{G}eneric- to \underline{I}ndividual-Language \underline{F}inetuning
}

\author{Tianze Shi\\
  Cornell University\\
  \texttt{tianze@cs.cornell.edu}\\\And
  Lillian Lee\\
  Cornell University\\
  \texttt{llee@cs.cornell.edu} \\
}

\date{}

\newtoggle{comment}
\toggletrue{comment}
\mathchardef\mhyphen="2D

\def\Snospace~{\S{}}

\newtheorem*{theorem*}{Theorem}

\newtheorem*{lemma*}{Lemma}

\newcommand{\posscite}[1]{\citeauthor{#1}'s \citeyearpar{#1}}

\newcommand{\concept}[1]{{\em #1}}

\newcommand{\bivec}[1]{\ensuremath{\mathbf{#1}}}

\newcommand{\vvec}[1]{\ensuremath{\bivec{v}_{#1}}}
\newcommand{\xvec}[1]{\ensuremath{\bivec{x}_{#1}}}

\newcommand{\deprel}[1]{\textsf{#1}}

\newcommand{\resnumber}[1]{\ensuremath{#1}}
\newcommand{\best}[1]{\ensuremath{\resnumber{\mathbf{#1}}}}

\newcommand{\xlmr}{\texttt{XLM-R}\xspace}
\newcommand{\cls}{\texttt{<s>}\xspace}
\newcommand{\sep}{\texttt{</s>}\xspace}

\begin{document}
\maketitle

\begin{abstract}

We present our contribution to the IWPT 2021 shared task
on parsing into enhanced Universal Dependencies.
Our main system component is a hybrid tree-graph parser that
integrates (a) predictions of
 spanning trees for the enhanced graphs with (b) additional graph edges not
 present in the spanning trees.
We also adopt a finetuning strategy where we first train a language-generic parser
on the concatenation of
data
from all available languages,
and then, in a second step, finetune on each individual language separately.
Additionally, we develop our own complete set of pre-processing
modules relevant to the shared task,
including tokenization, sentence segmentation,
and multi-word token expansion,
based on
pre-trained \xlmr models and
our own pre-training of character-level language models.
Our submission reaches a macro-average ELAS of $89.24$ on the test set.
It ranks top among all teams,
with a margin of more than $2$ absolute ELAS over
the next best-performing submission,
and best score on $16$ out of $17$ languages.
\end{abstract}

\section{Introduction}
\label{sec:intro}

The Universal Dependencies \citep[UD;][]{nivre+16,nivre+20} initiative
aims to provide cross-linguistically consistent annotations
for dependency-based syntactic analysis, and includes a large collection
of treebanks ($202$ for $114$ languages in UD 2.8).
Progress on the UD parsing problem has been steady \citep{zeman+17,zeman+18}, but existing approaches mostly focus on parsing into \concept{basic} UD trees,
where bilexical dependency relations among surface words must form single-rooted trees.
While these trees indeed contain rich syntactic information,
the adherence to tree representations can be insufficient
for certain
constructions
including coordination, gapping, relative clauses, and argument sharing through control and raising \citep{schuster-manning16}.

The IWPT 2020 \citep{bouma+20} and 2021 \citep{bouma+21}
shared tasks
focus on
parsing into \concept{enhanced} UD format,
where the
representation is connected graphs, rather than rooted trees.
The extension from trees to graphs allows direct treatment of a wider range of syntactic phenomena,
but it also poses a research challenge:
how to design
parsers
suitable for such enhanced UD graphs.

To address this setting,
we propose to use a tree-graph hybrid parser
leveraging the following key observation: since an enhanced UD
graph must be connected, it
must contain a spanning tree as a sub-graph.
These spanning trees may differ from basic UD trees,
but still allow us to use existing techniques
developed for dependency parsing,
including applying algorithms for finding maximum spanning trees to serve as
accurate global decoders.
Any additional dependency relations in the enhanced graphs not appearing
in the spanning trees
are then predicted on a per-edge basis.
We find that this tree-graph hybrid approach results in more accurate predictions
compared to a dependency graph parser that is combined with postprocessing steps to fix any graph connectivity issues.

Besides the enhanced graphs,
the
shared task setting poses two additional challenges.
Firstly, the evaluation is on $17$ languages from $4$ language families,
and not all the languages have large collections of annotated data:
the lowest-resource language, Tamil,
contains merely
$400$ training sentences
--- more than two magnitudes smaller than what is available for Czech.
To facilitate knowledge
sharing
between high-resource and low-resource languages,
we develop a two-stage finetuning strategy:
we first train a language-generic model
on the concatenation of all available training treebanks
from all languages provided by the shared task,
and then finetune on each language individually.

Secondly, the shared task demands parsing from raw text.
This requires accurate text processing pipelines
including modules for tokenization, sentence splitting,
and multi-word token expansion, in addition to enhanced UD parsing.
We build our own models for all these components; notably, we pre-train character-level masked language models on Wikipedia data, leading to
improvements on tokenization, the first component in the text processing pipeline.
Our multi-word token expanders combine the strengths of pre-trained learning-based models and rule-based approaches,
and achieve robust results, especially on low-resource languages.

Our system submission integrates the aforementioned solutions to the three main challenges
given by the shared task,
and ranks top among all submissions, with a macro-average EULAS of $90.16$ and ELAS of $89.24$.
Our system gives the best evaluation scores on all languages except for Arabic,
and has large margins (more than $5$ absolute ELAS) over the second-best systems
on Tamil and Lithuanian,
which are among languages with the smallest training treebanks.

\section{TGIF: Tree-Graph Integrated-Format Parser for Enhanced UD}
\label{sec:parser}

\subsection{Tree and Graph Representations for Enhanced UD}
The basic syntactic layer in UD is a single-rooted labeled dependency tree for each sentence,
whereas the enhanced UD layer
only requires that the set of dependency edges for each sentence form a connected graph.
In these connected graphs,
each word may have multiple parents,
there may be multiple roots for a sentence,
and the graphs may contain cycles,
but there must exist one path from at least one of the roots to each node.\footnote{
Enhanced UD graphs additionally allow insertion of phonologically-empty nodes
to recover elided elements in gapping constructions.
This is currently beyond the scope our system
and we use pre- and post-processing collapsing steps to handle empty nodes (\autoref{sec:notes}).
}

Accompanying the increase in expressiveness of the enhanced UD representation is
the challenge to produce structures that correctly satisfy graph-connectivity constraints during model inference.
We summarize the existing solutions proposed for the previous run of the shared task at IWPT 2020 \citep{bouma+20}
into four main categories:

\noindent
$\bullet$
\emph{Tree-based}: since the overlap between the enhanced UD graphs and the basic UD trees are typically significant,
and any deviations tend to be localized and tied to one of several certain syntactic constructions (e.g, argument sharing in a control structure),
one can repurpose tree-based parsers for producing enhanced UD graphs.
This category of approaches include
packing the additional edges from an enhanced graph into
the basic tree \citep{kanerva+20}
and using either rule-based or learning-based approaches
to convert a basic UD tree into an enhanced UD graph \citep{heinecke20,dehouck+20,attardi+20,ek-bernardy20}.\footnote{
The same idea has also been applied to the task
of conjunction propagation prediction \citep[e.g.,][]{grunewald+21}.
}

\noindent
$\bullet$
\emph{Graph-based}: alternatively,
one can
directly focus on the enhanced UD graph
with a semantic dependency graph parser
that predicts the existence and label of each candidate dependency edge.
But there is generally no guarantee that
the set of predicted edges will
form
a connected graph,
so a post-processing step is typically employed
to fix any connectivity issues.
This category of approaches includes the work of
\citet{wang+20a}, \citet{barry+20}, and \citet{grunewald-friedrich20}.\footnote{
\posscite{barry+20} parsers use basic UD trees as features,
but the output space is not restricted by the basic trees.
}

\noindent
$\bullet$
\emph{Transition-based}: \citet{hershcovich+20}
adapt a transition-based
solution.
Their system explicitly handles empty nodes through a specialized transition for inserting them;
it relies on additional post-processing  to ensure connectivity.

\noindent
$\bullet$
\emph{Tree-Graph Integrated}: \citet{he-choi20}
integrate
a tree parser and a graph parser,\footnote{
\citet{he-choi20}
describe their combo as an ``ensemble''
but we prefer the term ``integration'' for both their method and ours
(which is inspired by theirs),
since the two components are not, strictly speaking, targeting  same structures.
}
where the tree parser
produces
the basic UD tree,
and the graph parser predicts any additional edges.
During inference, all nodes are automatically connected through the tree parser,
and the graph parser allows flexibility in producing graph structures.\footnote{
The main difference from the tree-based approaches is that
the search space for additional graph edges
is unaffected by the predictions of basic UD trees
in an integrated approach.
}

The tree-based approaches are prone to error propagation,
since the predictions of the enhanced layer
rely heavily on the accuracy of basic UD tree parsing.
The graph-based and transition-based approaches
natively produce graph structures,
but they require post-processing to ensure connectivity.
Our system is a tree-graph integrated-format parser
that combines the strengths of the available global inference algorithms for tree parsing
and the flexibility of a graph parser,
without the need to use post-processing to fix connectivity issues.

\begin{figure}[t]
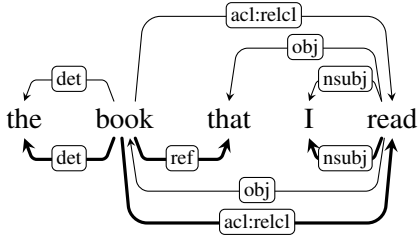

\centering
    \hspace*{-10pt}
\begin{dependency}
\begin{deptext}[column sep=15pt]
    the \& book \& that \& I \& read\\
\end{deptext}
    \depedge[edge height=8pt]{2}{1}{det}
    \depedge[edge height=32pt]{2}{5}{acl:relcl}
    \depedge[edge height=20pt]{5}{3}{obj}
    \depedge[edge height=8pt]{5}{4}{nsubj}
    \depedge[edge below, edge height=8pt, edge style=very thick]{2}{1}{det}
    \depedge[edge below, edge height=8pt, edge style=very thick]{5}{4}{nsubj}
    \depedge[edge below, edge height=8pt, edge style=very thick]{2}{3}{ref}
    \depedge[edge below, edge height=20pt, edge end x offset=2pt, edge start x offset=1pt]{5}{2}{obj}
    \depedge[edge below, edge height=32pt, edge style=very thick, edge start x offset=-5pt]{2}{5}{acl:relcl}
\end{dependency}
\caption{
An example with basic UD and enhanced UD annotations above and below the text respectively.
The extracted spanning tree (\autoref{sec:spanning-tree}) is bolded
and is different from the basic UD tree.
}
\label{fig:spanning-tree}

\end{figure}

\subsection{Spanning Tree Extraction}
\label{sec:spanning-tree}
A connected graph must contain a spanning tree,
and conversely,
if we first predict a spanning tree over all nodes,
and subsequently add additional edges,
then the resulting graph remains connected.
Indeed, this property is leveraged in some previously-proposed connectivity post-processing steps \citep[e.g.,][]{wang+20a},
but extracting a spanning tree based on scores from graph-prediction models creates a mismatch between training and inference.
\citet{he-choi20} instead train tree parsers and graph parsers separately and combine their prediction during inference,
but their tree parsers are trained on basic UD trees
whose edges are not always present in the enhanced UD layer.

Our solution refines \posscite{he-choi20} approach: we train tree parsers to predict spanning trees extracted from the enhanced UD graphs,
instead of basic UD trees,
to minimize train-test mismatch.
See \autoref{fig:spanning-tree} for an example.
Spanning tree extraction is in essence assignment of unique head nodes to all nodes in a graph,
subject to tree constraints.
For consistent extraction,
we apply the following rules:
\noindent
$\bullet$
If a node has a unique head in the enhanced graph,
there is no ambiguity in head assignment.

\noindent
$\bullet$
If a basic UD edge is present among the set of incoming edges to a given node,
include that basic UD edge in the spanning tree.

\noindent
$\bullet$
Otherwise, there must be multiple incoming edges, none of which are present in the basic UD tree.
We pick the parent node that is the ``highest'',
i.e., the closest to the root node,
in the basic tree.
The above head assignment steps
do not formally guarantee that the extracted structures will be trees,
but empirically,
we observe that the extraction results are indeed trees for
all training sentences.\footnote{Dear Reviewer 1: your question here in the submitted paper caused us to uncover a bug! Fixing it rectified the 4 training sentences that weren't originally getting trees.
}

\subsection{Parameterization}
Our parser architecture is adapted from that of \citet{dozat-manning17,dozat-manning18},
which forms the basis for the prior graph-based approaches in the IWPT 2020 shared task.
We predict unlabeled edges and labels separately,
and for the unlabeled edges,
we use a combination of a tree parser and a graph-edge prediction module.

\paragraph{Representation}
The first step is to extract contextual representations.
For this purpose, we use the pre-trained \xlmr model \citep{conneau+20},
which is trained on multilingual CommonCrawl data
and supports all $17$ languages in the shared task.
The \xlmr feature extractor is finetuned along with model training.
Given a length-$n$ input sentence $x=x_1,\ldots,x_n$
and layer $l$, we extract
\begin{equation*}
[\xvec{0}^l,\xvec{1}^l,\ldots,\xvec{n}^l]=\xlmr^l(\cls,x_1,\ldots,x_n,\sep),
\end{equation*}
where inputs to the \xlmr model are a concatenated sequence of word pieces from each UD word,
we denote the layer-$l$ vector corresponding to the last word piece in the word $x_i$ as $\xvec{i}^l$,
and the dummy root representations $\xvec{0}$s are taken from the special \cls token at the beginning of the sequence.

\paragraph{Deep Biaffine Function}
All our parsing components use deep biaffine functions (DBFs),
which score the interactions between pairs of words:
\begin{align*}
\text{DBF}(i,j)=&\vvec{i}^{\text{head}\top} U \vvec{j}^\text{mod}
+\bivec{b}^{\text{head}}\cdot \vvec{i}^{\text{head}}\\
&+\bivec{b}^{\text{mod}}\cdot \vvec{j}^{\text{mod}}
+b,
\end{align*}
where $\vvec{i}^\text{head}$ and $\vvec{j}^\text{mod}$
are non-linearly transformed vectors
from weighted average \xlmr vectors across different layers:
\begin{align*}
\vvec{i}^\text{head}=\text{ReLU}\left(W^\text{head}\sum\nolimits_l{\frac{e^{\alpha_l^\text{head}}}{\sum\nolimits_{l'} e^{\alpha_{l'}^\text{head}}}\xvec{i}^l}\right),
\end{align*}
and $\vvec{j}^\text{mod}$ is defined similarly.
Each DBF has its own trainable
weight matrices $U$, $W^\text{head}$, and $W^\text{mod}$, vectors $\bivec{b}^\text{head}$ and $\bivec{b}^\text{mod}$,
and scalars $b$, $\{\alpha_l^\text{head}\}$ and $\{\alpha_l^\text{mod}\}$.

\paragraph{Tree Parser}
To estimate the probabilities of head attachment for
each token $w_j$,
we define
\begin{equation*}
P(\text{head}(w_j)=w_i)=\text{softmax}_i(\text{DBF}^\text{tree}(i,j)).
\end{equation*}
The tree parsing models are trained with cross-entropy loss,
and we use a non-projective maximum spanning tree algorithm \citep{chu-liu65,edmonds67}
for global inference.

\paragraph{Graph Parser}
In addition to the spanning trees,
we make independent predictions on
the existence of any extra edges in the enhanced UD graphs by
\begin{equation*}
P(\exists \text{edge}~w_i\rightarrow w_j)=\text{sigmoid}(\text{DBF}^\text{graph}(i,j)).
\end{equation*}
We train the graph parsing model with
a cross entropy objective,
and during inference,
any edges with probabilities $\ge 0.5$ are included in the outputs.

\paragraph{Relation Labeler}
For each edge in the unlabeled graph,
we predict the relation label via
\begin{equation*}
P(\text{lbl}(w_i \rightarrow w_j)=r)=\text{softmax}_r(\text{DBF}^{\text{rel-}r}(i,j)),
\end{equation*}
where we have as many deep biaffine functions
as the number of candidate relation labels in the data.
To reduce the large number of potential labels due to lexicalization,
the relation labeler operates on a de-lexicalized version of the labels,
and then a re-lexicalization step expands the predicted labels into their full forms (\autoref{sec:notes}).

\paragraph{Training}
The above three components are separately parameterized,
and during training,
we optimize for the sum of their
corresponding cross-entropy loss functions.

\subsection{Empirical Comparisons}

\begin{table}[t]
\small
\centering
\begin{tabular}{l@{\hspace{4pt}}c@{\hspace{4pt}}c@{\hspace{4pt}}c@{\hspace{4pt}}c}
\toprule
\multirow{2}{*}{Language} & \multicolumn{2}{c}{Direct Training}  & \multirow{2}{*}{Generic} & \multirow{2}{*}{Finetuned} \\
& Graph+Fix & Tree-Graph & & \\
\midrule
Arabic & \resnumber{80.34} & \resnumber{80.30} & \resnumber{80.57} &\best{80.63} \\
Bulgarian & \resnumber{91.81} & \resnumber{92.00} & \resnumber{91.69} &\best{92.30} \\
Czech & \resnumber{92.93} & \best{92.98} & \resnumber{92.94} &\best{92.98} \\
Dutch & \resnumber{92.14} & \resnumber{92.13} & \resnumber{92.03} &\best{92.21} \\
English & \resnumber{88.38} & \resnumber{88.51} & \resnumber{88.44} &\best{88.83} \\
Estonian & \best{89.53} & \resnumber{89.42} & \resnumber{89.22} &\resnumber{89.40} \\
Finnish & \resnumber{91.97} & \resnumber{92.10} & \resnumber{91.84} &\best{92.48} \\
French & \resnumber{94.46} & \resnumber{94.51} & \resnumber{94.26} &\best{95.52} \\
Italian & \resnumber{93.04} & \resnumber{93.24} & \resnumber{93.26} &\best{93.41} \\
Latvian & \resnumber{88.47} & \resnumber{88.42} & \resnumber{88.38} &\best{89.78} \\
Lithuanian & \resnumber{90.57} & \resnumber{90.63} & \resnumber{90.47} &\best{90.85} \\
Polish & \resnumber{91.28} & \resnumber{91.48} & \resnumber{91.28} &\best{91.63} \\
Russian & \resnumber{93.47} & \best{93.50} & \resnumber{93.37} &\resnumber{93.47} \\
Slovak & \resnumber{93.70} & \resnumber{93.83} & \resnumber{94.00} &\best{95.44} \\
Swedish & \resnumber{90.35} & \resnumber{90.48} & \resnumber{90.33} &\best{91.57} \\
Tamil & \resnumber{66.24} & \resnumber{66.82} & \resnumber{67.35} &\best{68.95} \\
Ukrainian & \resnumber{92.98} & \resnumber{92.94} & \resnumber{93.24} &\best{93.89} \\
\midrule
Average & \resnumber{89.51} & \resnumber{89.61} & \resnumber{89.57} &\best{90.20} \\

\bottomrule
\end{tabular}
\caption{
Dev-set ELAS (\%) results,
comparing graph parsers with connectivity-fixing postprocessing
against tree-graph integrated models (\autoref{sec:parser})
and comparing parsers trained directly on each language,
generic-language parsers,
and parsers finetuned on individual languages
from the generic-language checkpoint (\autoref{sec:finetune}).
}
\label{tbl:dev}

\end{table}

In \autoref{tbl:dev},
we compare our tree-graph integrated-format parser
with a fully graph-based approach.
The graph-based baseline uses the same feature extractor,
graph parser, and relation labeler modules,
but it omits the tree parser for producing spanning trees,
and we apply post-processing steps to ensure connectivity of the output graphs.
Our tree-graph integrated-format parser outperforms
the graph-based baseline on $12$ out of the $17$ test languages (binomial test, $p=0.07$).

\section{TGIF: Two-Stage Generic- to Individual-Language Finetuning}
\label{sec:finetune}

In addition to the tree-graph integration approach,
our system submission also features a two-stage finetuning strategy.
We first train a language-generic model on the concatenation of all available training treebanks in the shared task data regardless of their source languages,
and then finetune on each individual language in a second step.

This two-stage finetuning strategy is designed to
encourage knowledge sharing across different languages,
especially from high-resource languages to lower-resource ones.
In our experiment results as reported in \autoref{tbl:dev},
we find that this strategy is indeed beneficial for
the majority of languages, especially those with small training corpora (e.g., $2.13$ and $1.01$ absolute ELAS improvements on Tamil and French respectively),
though this comes at the price of slightly decreased accuracies on
high-resource languages (e.g., $-0.02$ on Estonian and $-0.03$ on Russian).
Additionally,
we find that the language-generic model
achieves reasonably competitive performance
when compared with the set of models directly trained on each individual language.
This suggests that
practitioners may opt to use a single model for parsing all languages
if there is a need to lower disk and memory footprints,
without much loss in accuracy.

\section{Pre-TGIF: Pre-Training Grants Improvements Full-Stack}
\label{sec:preprocess}

Inspired by the recent success of pre-trained language models on a wide range of NLP tasks \citep[][\emph{inter alia}]{peters+18,devlin+19,conneau+20},
we build our own text processing pipeline based on pre-trained language models.
Due to limited time and resources,
we only focus on components relevant to the shared task,
which include tokenization, sentence splitting, and multi-word token (MWT) expansion.

\subsection{Tokenizers with Character-Level Masked Language Model Pre-Training}

We follow  state-of-the-art strategies \citep{qi+20,nguyen+21} for tokenization
and model the task as a tagging problem on sequences of characters.
But in contrast to prior methods
where tokenization and sentence segmentation
are bundled into the same prediction stage,
we tackle tokenization in isolation,
and for each character,
we make a binary prediction as to whether a token ends
at the current character position or not.

An innovation in our tokenization is that
we finetune character-based language models trained on Wikipedia data.
In contrast, existing approaches typically
use randomly-initialized models \citep{qi+20}
or use pre-trained models on subword units instead of characters \citep{nguyen+21}.

We follow \citet{devlin+19} and pre-train
our character-level sequence models
using a masked language modeling objective:
during training, we randomly replace $15\%$ of the characters with a special mask symbol
and the models are trained to predict the identity of those characters in the original texts.
Due to computational resource constraints,
we adopt a small-sized architecture based on simple recurrent units \citep{lei+18}.\footnote{
Simple recurrent units are a fast variant of recurrent neural networks.
In our preliminary experiments,
they result in lower accuracies than long-short term memory networks (LSTMs),
but are $2$-$5$ times faster,
depending on sequence lengths.
}
We pre-train our models on Wikipedia data\footnote{
We extract Wikipedia texts using WikiExtractor \citep{attardi15}
from Wikipedia dumps dated 2021-04-01.
}
and each model takes roughly $2$ days to complete $500$k optimization steps on a single GTX 2080Ti GPU.

\begin{table*}[t]
\small
\centering
\begin{tabular}{l|ccc|ccc|ccc}
\toprule
\multirow{2}{*}{Treebank} & \multicolumn{3}{c|}{Token} & \multicolumn{3}{c|}{Sentence} & \multicolumn{3}{c}{Word} \\
 & Stanza & Trankit & Ours & Stanza & Trankit & Ours & Stanza & Trankit & Ours \\
\midrule

Arabic-PADT &
\resnumber{99.98} & \resnumber{99.95} & \best{99.99} &
\resnumber{80.43} & \resnumber{96.79} & \best{96.87} &
\resnumber{97.88} & \best{99.39} & \resnumber{98.70} \\

Bulgarian-BTB &
\resnumber{99.93} & \resnumber{99.78} & \best{99.95} &
\resnumber{97.27} & \resnumber{98.79} & \best{99.06} &
\resnumber{99.93} & \resnumber{99.78} & \best{99.95} \\

Czech-FicTree &
\resnumber{99.97} & \best{99.98} & \resnumber{99.97} &
\resnumber{98.60} & \resnumber{99.50} & \best{99.54} &
\resnumber{99.96} & \best{99.98} & \resnumber{99.96} \\

Czech-CAC &
\best{99.99} & \best{99.99} & \best{99.99} &
\best{100.00} & \best{100.00} & \best{100.00} &
\resnumber{99.97} & \resnumber{99.98} & \best{99.99} \\

Dutch-Alpino &
\best{99.96} & \resnumber{99.43} & \resnumber{99.86} &
\resnumber{89.98} & \resnumber{90.65} & \best{94.45} &
\best{99.96} & \resnumber{99.43} & \resnumber{99.86} \\

Dutch-LassySmall &
\resnumber{99.90} & \resnumber{99.36} & \best{99.94} &
\resnumber{77.95} & \resnumber{92.60} & \best{94.23} &
\resnumber{99.90} & \resnumber{99.36} & \best{99.94} \\

English-EWT &
\best{99.01} & \resnumber{98.67} & \resnumber{98.79} &
\resnumber{81.13} & \resnumber{90.49} & \best{92.70} &
\best{99.01} & \resnumber{98.67} & \resnumber{98.79} \\

English-GUM &
\best{99.82} & \resnumber{99.52} & \resnumber{98.88} &
\resnumber{86.35} & \resnumber{91.60} & \best{95.11} &
\best{99.82} & \resnumber{99.52} & \resnumber{99.18} \\

Estonian-EDT &
\best{99.96} & \resnumber{99.75} & \resnumber{99.95} &
\resnumber{93.32} & \resnumber{96.58} & \best{96.60} &
\best{99.96} & \resnumber{99.75} & \resnumber{99.95} \\

Estonian-EWT &
\best{99.20} & \resnumber{97.76} & \resnumber{98.72} &
\resnumber{67.14} & \resnumber{82.58} & \best{89.37} &
\best{99.20} & \resnumber{97.76} & \resnumber{98.72} \\

Finnish-TDT &
\best{99.77} & \resnumber{99.71} & \resnumber{99.76} &
\resnumber{93.05} & \resnumber{97.22} & \best{98.26} &
\best{99.73} & \resnumber{99.72} & \best{99.73} \\

French-Sequoia &
\best{99.90} & \resnumber{99.81} & \resnumber{99.88} &
\resnumber{88.79} & \resnumber{94.07} & \best{96.82} &
\resnumber{99.58} & \resnumber{99.78} & \best{99.84} \\

Italian-ISDT &
\best{99.91} & \resnumber{99.88} & \resnumber{99.90} &
\resnumber{98.76} & \best{99.07} & \best{99.07} &
\resnumber{99.76} & \best{99.86} & \resnumber{99.83} \\

Latvian-LVTB &
\best{99.82} & \resnumber{99.73} & \resnumber{99.80} &
\resnumber{99.01} & \resnumber{98.69} & \best{99.26} &
\best{99.82} & \resnumber{99.73} & \resnumber{99.80} \\

Lithuanian-ALKSNIS &
\resnumber{99.87} & \resnumber{99.84} & \best{99.99} &
\resnumber{88.79} & \resnumber{95.72} & \best{96.22} &
\resnumber{99.87} & \resnumber{99.84} & \best{99.99} \\

Polish-LFG &
\best{99.95} & \resnumber{98.34} & \resnumber{99.84} &
\resnumber{99.83} & \resnumber{99.57} & \best{99.88} &
\best{99.95} & \resnumber{98.34} & \resnumber{99.89} \\

Polish-PDB &
\resnumber{99.87} & \best{99.93} & \resnumber{99.49} &
\resnumber{98.39} & \resnumber{98.71} & \best{99.66} &
\resnumber{99.83} & \best{99.92} & \resnumber{99.84} \\

Russian-SynTagRus &
\resnumber{99.57} & \resnumber{99.71} & \best{99.73} &
\resnumber{98.86} & \resnumber{99.45} & \best{99.54} &
\resnumber{99.57} & \resnumber{99.71} & \best{99.73} \\

Slovak-SNK &
\best{99.97} & \resnumber{99.94} & \resnumber{99.95} &
\resnumber{90.93} & \best{98.49} & \resnumber{96.72} &
\best{99.97} & \resnumber{99.94} & \resnumber{99.94} \\

Swedish-Talbanken &
\best{99.97} & \resnumber{99.91} & \best{99.97} &
\resnumber{98.85} & \resnumber{99.26} & \best{99.34} &
\best{99.97} & \resnumber{99.91} & \best{99.97} \\

Tamil-TTB &
\resnumber{99.58} & \resnumber{98.33} & \best{99.63} &
\resnumber{95.08} & \best{100.00} & \best{100.00} &
\resnumber{91.42} & \resnumber{94.44} & \best{95.34} \\

Ukrainian-IU &
\resnumber{99.81} & \resnumber{99.77} & \best{99.86} &
\resnumber{96.65} & \resnumber{97.55} & \best{98.38} &
\resnumber{99.79} & \resnumber{99.76} & \best{99.84} \\

\bottomrule
\end{tabular}
\caption{
Test-set F1 scores for tokenization, sentence segmentation,
and MWT expansion,
comparing Stanza \citep{qi+20}, Trankit \citep{nguyen+21},
and our system submission.
Our system results are from the shared task official evaluations; Stanza and Trankit results are reported in the Trankit documentation
with models trained on UD 2.5.
\emph{Caveat}: the results may not be strictly comparable due to treebank version mismatch.
}

\label{tbl:pipeline-results}

\end{table*}

\subsection{Sentence Splitters}
We split texts into sentences from sequences of tokens
instead of characters \citep{qi+20}.
Our approach resembles that of \citet{nguyen+21}.\footnote{
An important difference is that
our sentence splitters are aware of token boundaries
and the models are restricted from making token-internal
sentence splitting decisions.
}
This allows our models to condense information from a wider range of contexts
while still reading the same number of input symbols.
The sentence splitters are trained to make binary predictions
at each token position
on whether a sentence ends there.
We adopt the same two-stage finetuning strategy as for our parsing modules based on pre-trained \xlmr feature extractors (\autoref{sec:finetune}).

\subsection{Multi-Word Token (MWT) Expanders}

The UD annotations distinguish between tokens and words.
A word corresponds to a consecutive sequence of characters in the surface raw text
and may contain one or more syntactically-functioning words.
We break down the MWT expansion task into
first deciding whether or not to expand a given token
and then performing the actual expansion.
For the former,
we train models to make a binary prediction on each token,
and we use pre-trained \xlmr models as our feature extractors.

For the MWT expansion step once the tokens are identified through our classifiers,
we use a combination of lexicon- and rule-based approaches.
If the token form is seen in the training data,
we adopt the most frequently used way to split it into multiple words.
Otherwise, we invoke a set of language-specific handwritten rules developed from and tuned on the training data;
a typical rule iteratively splits off an identified prefix or suffix from the remainder of the token.

\subsection{Lemmatizers}

While the shared task requires lemmatized forms
for constructing the lexicalized enhanced UD labels,
we only need to predict lemmas for a small percentage of words.
Empirically, these words tend to be function words and have a unique lemma per word type.
Thus, we use a full lexicon-based approach to (incomplete) lemmatization.
Whenever a lemma is needed during the label re-lexicalization step,
we look the word up in a dictionary extracted from the training data.

\subsection{Evaluation}

We compare our text-processing pipeline components
with two state-of-the-art toolkits, Stanza \citep{qi+20} and Trankit \citep{nguyen+21} in \autoref{tbl:pipeline-results}.
We train our models per-language instead of per-treebank
to accommodate the shared task setting,
so our models are at a disadvantage
when there are multiple training treebanks for a language
that have different tokenization/sentence splitting conventions (e.g., English-EWT and English-GUM handle word contractions differently).
Despite this,
our models are highly competitive in terms of
tokenization and MWT expansion,
and we achieve significantly better sentence segmentation results
across most treebanks.
We hypothesize that a sequence-to-sequence MWT expansion approach, similar to the ones underlying Stanza and Trankit,
may provide further gains
to morphologically-rich languages
that cannot be sufficiently modeled via handwritten rules,
notably Arabic.

\section{Other Technical Notes}
\label{sec:notes}

\paragraph{Hyperparameters}

We report our hyperparameters in the Appendix.

\paragraph{Empty nodes}
Enhanced UD graphs may contain empty nodes
in addition to the words in the surface form.
Our parser does not support empty nodes,
so we follow the official evaluation practice
and collapse  relation paths with empty nodes into
composite relations during training and inference.

\paragraph{Multiple relations}
In some cases, there can be multiple relations between the same pair of words.
We follow \citet{wang+20a} and merge all these relations into a composite label,
and re-expand them during inference.

\begin{table*}[t]
\small
\centering
\begin{tabular}{lccccccccc}
\toprule
Language & combo & dcu\_epfl & fastparse & grew & nuig & robertnlp & shanghaitech & tgif (Ours) & unipi \\
\midrule
Arabic & \resnumber{76.39} & \resnumber{71.01} & \resnumber{53.74} & \resnumber{71.13} & -- & \resnumber{81.58} & \best{82.26} & \resnumber{81.23} & \resnumber{77.13} \\

Bulgarian & \resnumber{86.67} & \resnumber{92.44} & \resnumber{78.73} & \resnumber{88.83} & \resnumber{78.45} & \resnumber{93.16} & \resnumber{92.52} & \best{93.63} & \resnumber{90.84} \\

Czech & \resnumber{89.08} & \resnumber{89.93} & \resnumber{72.85} & \resnumber{87.66} & -- & \resnumber{90.21} & \resnumber{91.78} & \best{92.24} & \resnumber{88.73} \\

Dutch & \resnumber{87.07} & \resnumber{81.89} & \resnumber{68.89} & \resnumber{84.09} & -- & \resnumber{88.37} & \resnumber{88.64} & \best{91.78} & \resnumber{84.14} \\

English & \resnumber{84.09} & \resnumber{85.70} & \resnumber{73.00} & \resnumber{85.49} & \resnumber{65.40} & \resnumber{87.88} & \resnumber{87.27} & \best{88.19} & \resnumber{87.11} \\

Estonian & \resnumber{84.02} & \resnumber{84.35} & \resnumber{60.05} & \resnumber{78.19} & \resnumber{54.03} & \resnumber{86.55} & \resnumber{86.66} & \best{88.38} & \resnumber{81.27} \\

Finnish & \resnumber{87.28} & \resnumber{89.02} & \resnumber{57.71} & \resnumber{85.20} & -- & \resnumber{91.01} & \resnumber{90.81} & \best{91.75} & \resnumber{89.62} \\

French & \resnumber{87.32} & \resnumber{86.68} & \resnumber{73.18} & \resnumber{83.33} & -- & \resnumber{88.51} & \resnumber{88.40} & \best{91.63} & \resnumber{87.43} \\

Italian & \resnumber{90.40} & \resnumber{92.41} & \resnumber{78.32} & \resnumber{90.98} & -- & \resnumber{93.28} & \resnumber{92.88} & \best{93.31} & \resnumber{91.81} \\

Latvian & \resnumber{84.57} & \resnumber{86.96} & \resnumber{66.43} & \resnumber{77.45} & \resnumber{56.67} & \resnumber{88.82} & \resnumber{89.17} & \best{90.23} & \resnumber{83.01} \\

Lithuanian & \resnumber{79.75} & \resnumber{78.04} & \resnumber{48.27} & \resnumber{74.62} & \resnumber{59.13} & \resnumber{80.76} & \resnumber{80.87} & \best{86.06} & \resnumber{71.31} \\

Polish & \resnumber{87.65} & \resnumber{89.17} & \resnumber{71.52} & \resnumber{78.20} & -- & \resnumber{89.78} & \resnumber{90.66} & \best{91.46} & \resnumber{88.31} \\

Russian & \resnumber{90.73} & \resnumber{92.83} & \resnumber{78.56} & \resnumber{90.56} & \resnumber{66.33} & \resnumber{92.64} & \resnumber{93.59} & \best{94.01} & \resnumber{90.90} \\

Slovak & \resnumber{87.04} & \resnumber{89.59} & \resnumber{64.28} & \resnumber{86.92} & \resnumber{67.45} & \resnumber{89.66} & \resnumber{90.25} & \best{94.96} & \resnumber{86.05} \\

Swedish & \resnumber{83.20} & \resnumber{85.20} & \resnumber{67.26} & \resnumber{81.54} & \resnumber{63.12} & \resnumber{88.03} & \resnumber{86.62} & \best{89.90} & \resnumber{84.91} \\

Tamil & \resnumber{52.27} & \resnumber{39.32} & \resnumber{42.53} & \resnumber{58.69} & -- & \resnumber{59.33} & \resnumber{58.94} & \best{65.58} & \resnumber{51.73} \\

Ukrainian & \resnumber{86.92} & \resnumber{86.09} & \resnumber{63.42} & \resnumber{83.90} & -- & \resnumber{88.86} & \resnumber{88.94} & \best{92.78} & \resnumber{87.51} \\

\midrule
Average & \resnumber{83.79} & \resnumber{83.57} & \resnumber{65.81} & \resnumber{81.58} & -- & \resnumber{86.97} & \resnumber{87.07} & \best{89.24} & \resnumber{83.64} \\

Rank & $4$ & $6$ & $8$ & $7$ & $9$ & $3$ & $2$ & $1$ & $5$ \\

\bottomrule
\end{tabular}
\caption{
Official ELAS (\%) evaluation results.
Our submission ranks  first on $16$ out of the $17$ languages.
}

\label{tbl:official-eval}

\end{table*}

\paragraph{De-lexicalization and re-lexicalization}
Certain types of relation labels include lexicalized information,
resulting in a large relation label set.
For example, \deprel{nmod:in} contains a lemma ``in''
that is taken from the modifier with a \deprel{case} relation.
To combat this, we follow \posscite{grunewald-friedrich20} strategy
and replace the lemmas\footnote{
We find that using lemmas instead of word forms
significantly improves coverage of the lexicalized labels.
}
with placeholders consisting of
their corresponding relation labels.
The previous example would result in a de-lexicalized label of \deprel{nmod:[case]}.
During inference, we apply a re-lexicalization step
to reconstruct the original full relation labels given our predicted graphs.
We discard the lexicalized portions of the relation labels
when errors occur either in de-lexicalization (unable to locate the source child labels to match the lemmas)
or re-lexicalization (unable to find corresponding placeholder relations).

\paragraph{Sequence length limit}
Pre-trained language models typically have a limit
on their input sequence lengths.
The \xlmr model has a limit of $512$ word pieces.
For a small number of sentences longer than that,
we discard word-internal word pieces, i.e., keep a prefix and a suffix of word pieces, of the longest words
to fit within limit.

\paragraph{Multiple Treebanks Per Language}
Each language in the shared task can have one or more treebanks for training and/or testing.
During evaluation,
there is no explicit information regarding the source treebank of the piece of input text.
Instead of handpicking a training treebank for each language,
we simple train and validate on the concatenation of all available data for each language.

\paragraph{Training on a single GPU}
The \xlmr model has large number of parameters,
which makes it challenging to finetune on a single GPU.
We use a batch size of $1$ and accumulate gradients across multiple batches
to lower the usage of GPU RAM.
When this strategy alone is insufficient,
e.g., when training the language-generic model,
we additionally freeze the initial embedding layer of the model.

\section{Official Evaluation}
\label{sec:evaluation}

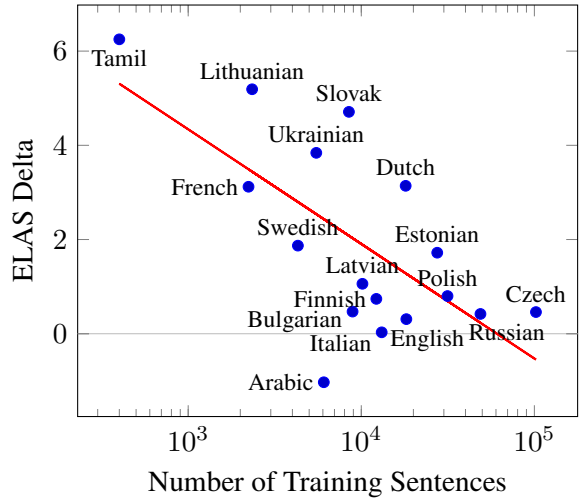
\begin{figure}[t]

    \begin{tikzpicture}
        \begin{axis}[
            width=0.51*\textwidth,
            compat=newest,
            xmode=log,
            visualization depends on = \thisrow{angle} \as \angle,
            enlargelimits = true,
            extra y ticks       = 0,
            extra y tick labels = ,
            extra y tick style  = { grid = major },
            xlabel = {Number of Training Sentences},
            ylabel = {ELAS Delta},
        ]
            \addplot+[
                mark=*, only marks,
                nodes near coords,
                point meta=explicit symbolic,
                every node near coord/.append style={black,anchor=\angle,font=\small},
            ] table [
                x=sents, y=delta,
                meta=lan,
            ] {fig-delta.dat};
            \addplot [no markers, thick, red] table [x=sents,y={create col/linear regression={y=delta}}] {fig-delta.dat};
        \end{axis}
        \end{tikzpicture}

\caption{
The per-language delta ELAS between our submission and the best performing system other than ours,
as a function of (the log of the) number of training sentences. (For Italian, the difference is quite small but still positive.)
Our models achieve larger improvements on lower-resource languages.
}
\label{fig:delta}

\end{figure}

The shared task performs evaluation on UD treebanks that have enhanced UD annotations across $17$ languages:
Arabic \citep{hajic+09a},
Bulgarian \citep{simov+04},
Czech \citep{hladka+10,bejcek+13,jelinek17},
Dutch \citep{vanderbeek+02,bouma-vannoord17},
English \citep{silveira+14,zeldes17},
Estonian \citep{muischnek+14,muischnek+19},
Finnish \citep{haverinen+14,pyysalo+15},
French \citep{candito+14,seddah-candito16},
Italian \citep{bosco+13},
Latvian \citep{pretkalnina+18a},
Lithuanian \citep{bielinskiene+16},
Polish \citep{patejuk-przepiorkowski18,wroblewska18},
Russian \citep{droganova+18},
Slovak \citep{zeman18},
Swedish \citep{nivre-megyesi07},
Tamil \citep{ramasamy-zabokrtsky12},
Ukrainian \citep{kotsyba+},
and multilingual parallel treebanks \citep{zeman+17}.

\autoref{tbl:official-eval}
shows the official ELAS evaluation results
of all $9$ participating systems in the shared task.\footnote{
Reproduced from \url{https://universaldependencies.org/iwpt21/results.html}.
}
Our system has the top performance
on $16$ out of $17$ languages,
and it is also the best in terms of macro-average across all languages.
On average, we outperform the second best system
by a margin of more than $2$ ELAS points in absolute terms,
or more than $15\%$ in relative error reduction.

\autoref{fig:delta} visualizes the ``delta ELAS'' between
our submission and the best result other than ours
on a per-language basis,
plotted against the training data size for each language.
Our system sees larger improvements on lower-resource languages,
where we have more than $5$-point leads on Tamil and Lithuanian,
two languages among those with the smallest number of training sentences.

\section{Closing Remarks}
\label{sec:conclusion}

Our submission to the IWPT 2021 shared task combines three main techniques:
(1) tree-graph integrated-format parsing (graph $\rightarrow$ spanning tree $\rightarrow$ additional edges)
(2) two-stage generic- to individual-language finetuning,
and (3) pre-processing pipelines powered by language model pre-training.
Each of the above contributes to our system performance positively,\footnote{Comparing the 3 components:  multilingual pre-training has a greater effect than the tree-graph parsing design.  Sentence segmentation performance (SSP) doesn't necessarily translate to ELAS, so our SSP's large relative improvement at SS doesn't imply that SS is the biggest contributor to our system.}
and by combining all three techniques,
our system
achieves the best ELAS results
on $16$ out of $17$ languages,
as well as top macro-average across all languages,
among all system submissions.
Additionally,
our system shows more relative strengths on
lower-resource languages.

Due to time and resource constraints,
our system adopts the same set of techniques across all languages
and we train a single set of models for our primary submission.
We leave it to future work to explore
language-specific methods and/or model combination and ensemble techniques
to further enhance model accuracies.

\paragraph{Acknowledgements}

We thank the anonymous reviewers for their constructive and detailed comments, and the task organizers for their flexibility regarding page limits.
This work was supported in part by a Bloomberg Data Science Ph.D. Fellowship to Tianze Shi and a gift from Bloomberg to Lillian Lee.

\bibliography{ref}
\bibliographystyle{acl_natbib}

\newpage

\appendix

\section{Hyperparameters}

{
    \small
\begin{tabular}{ll}
\toprule
\multicolumn{2}{c}{\textbf{Character-level Language Model Pre-training}}\\
\multicolumn{2}{l}{\textit{Optimization:}}\\
\quad Optimizer & RAdam \cite{liu+20a} \\
\quad Batch size & $128$ \\
\quad Number of steps & $500{,}000$ \\
\quad Initial learning rate & $3\times 10^{-4}$ \\
\quad Weight decay & $0.1$ \\
\quad Gradient clipping & $1.0$ \\
\multicolumn{2}{l}{\textit{Simple Recurrent Units:}}\\
\quad Sequence length limit & $512$ \\
\quad Vocab size & $512$ \\
\quad Embedding size & $256$ \\
\quad Hidden size & $256$ \\
\quad Numer of layers & $8$ \\
\quad Dropout & $0.3$ \\
\midrule
\multicolumn{2}{c}{\textbf{Tokenizer}}\\
\multicolumn{2}{l}{\textit{Optimization:}}\\
\quad Optimizer & RAdam \\
\quad Batch size & $32$ \\
\quad Initial learning rate & $5\times 10^{-5}$ \\
\quad Weight decay & $0$ \\
\quad Gradient clipping & $1.0$ \\
\multicolumn{2}{l}{\textit{Multi-layer Perceptrons (MLPs):}}\\
\quad Number of layers & $1$ \\
\quad Hidden size & $500$ \\
\quad Dropout & $0.5$ \\
\midrule
\multicolumn{2}{c}{\textbf{Sentence Splitter, MWT Expander, and Parser}}\\
\quad Pre-trained model & \xlmr (Large) \\
\multicolumn{2}{l}{\textit{Optimization:}}\\
\quad Optimizer & RAdam \\
\quad Batch size & $8$ \\
\quad Initial learning rate & $1\times 10^{-5}$ \\
\quad Second-stage learning rate & $1\times 10^{-6}$ \\
\quad Weight decay & $0$ \\
\quad Gradient clipping & $1.0$ \\
\multicolumn{2}{l}{\textit{Tagger MLPs (Sentence Splitter, MWT Expander):}}\\
\quad Number of layers & $1$ \\
\quad Hidden size & $400$ \\
\quad Dropout & $0.5$ \\
\multicolumn{2}{l}{\textit{Parser MLPs (Unlabeled Tree and Graph Parsers):}}\\
\quad Number of layers & $1$ \\
\quad Hidden size & $383$ \\
\quad Dropout & $0.33$ \\
\multicolumn{2}{l}{\textit{Parser MLPs (Relation Labeler):}}\\
\quad Number of layers & $1$ \\
\quad Hidden size & $255$ \\
\quad Dropout & $0.33$ \\

\bottomrule

\end{tabular}
}

\end{document}